\documentclass[journal]{IEEEtran}
\usepackage{subfig}
\usepackage{graphicx}
\usepackage{soul}
\usepackage[dvipsnames]{xcolor}
\usepackage{enumerate}
\usepackage{enumitem}

\usepackage[activate]{microtype}
\usepackage{hyperref}
\usepackage[acronym, shortcuts, nohypertypes={acronym}]{glossaries}
\usepackage[capitalize]{cleveref}

\usepackage{tablefootnote}
\usepackage[hang,flushmargin]{footmisc} % to remove indentation from footnotes
\usepackage{booktabs}

\newacronym{cl}{CL}{\textit{curriculum learning}}
\newacronym{ml}{ML}{\textit{machine learning}}
\newacronym{dnn}{DNN}{\textit{deep neural network}}
\newacronym{cnn}{CNN}{\textit{convolutional neural network}}
\newacronym{fcnn}{FCNN}{\textit{fully connected neural network}}
\newacronym{dl}{DL}{\textit{deep learning}}
\newacronym{cv}{CV}{\textit{computer vision}}
\newacronym{ca}{CA}{\textit{computer audition}}
\newacronym{asc}{ASC}{\textit{acoustic scene classification}}
\newacronym{snr}{SNR}{\textit{signal-to-noise-ratio}}
\newacronym{sf}{SF}{\textit{scoring function}}
\newacronym{pf}{PF}{\textit{pacing function}}
\newacronym{cumacc}{CumAcc}{\textit{cumulative accuracy}}
\newacronym{fit}{FIT}{\textit{first iteration}}
\newacronym{celoss}{CELoss}{\textit{cross-entropy-loss}}
\newacronym{tt}{TT}{\textit{transfer teacher}}
\newacronym{pd}{PD}{\textit{prediction depth}}
\newacronym{sd}{SD}{\textit{sample difficulty}}
\newacronym{knn}{KNN}{\textit{k-nearest neighbours}}
\newacronym{svc}{SVC}{\textit{support vector classifier}}
\newacronym{ser}{SER}{\textit{speech emotion recognition}}
\newacronym{mvt}{MVT}{\textit{minimum viewing time}}
\newacronym{cvloss}{CVLoss}{\textit{cross-validation loss}}
\newacronym{cvst}{CVST}{\textit{cross-validated self-thought}}
\newacronym{gap}{GAP}{\textit{global average pooling}}
\newacronym{spl}{SPL}{\textit{self-paced learning}}
\newacronym{asr}{ASR}{\textit{automatic speech recognition}}
\newacronym{resnet}{ResNet}{\textit{residual network}}
\newacronym{pann}{PANN}{\textit{large-scale pre-trained audio neural network}}
\newacronym{acl}{ACL}{\textit{anti-curriculum learning}}
\newacronym{rcl}{RCL}{\textit{random curriculum learning}}
\newacronym{nlp}{NLP}{\textit{natural language processing}}
\newacronym{pcc}{PCC}{\textit{Pearson correlation coefficient}}

\usepackage{xspace}
\newcommand{\eg}{e.\,g.,\xspace}
\newcommand{\ie}{i.\,e.,\xspace}
\newcommand{\wrt}{w.\,r.\,t.\xspace}
\newcommand{\cf}{cf.\xspace}
\newcommand{\sota}{state-of-the-art\xspace}
\newcommand{\etal}{et~al.\xspace}

\begin{document}

% \title{An Extensive Analysis of Curriculum Learning\\
% }

% \title{On the Role of Scoring Functions for Curriculum Learning\\
% }

\title{Does the Definition of Difficulty Matter?\\ Scoring Functions and their Role for \\Curriculum Learning
}
% \title{Does the Definition of Difficulty Matter?\\ An Analysis of Scoring Functions and their Role for Curriculum Learning
% }

% Does Sample Difficulty Matter? An Analysis of Scoring Functions and Their Impact on Curriculum Learning Performance

%The Role of Scoring Functions in Curriculum Learning 

%Does the Scoring Function Matter for Curriculum Learning?  

\author{
  \textbf{Simon Rampp}$^{1}$, \textbf{Manuel Milling}$^{1,2}$, \textbf{Andreas Triantafyllopoulos}$^{1,2}$, \textbf{Björn W. Schuller}$^{1,2,3,4}$ \\
  $^1$CHI -- Chair of Health Informatics, Technical University of Munich, Munich, Germany \\
  $^2$MCML -- Munich Center for Machine Learning, Munich, Germany \\
  $^3$GLAM -- Group on Language, Audio, \& Music, Imperial College, London, UK\\
  $^4$MDSI -- Munich Data Science Institute, Munich, Germany\\
  \texttt{\{simon.rampp;manuel.milling;andreas.triantafyllopoulos;schuller\}@tum.de}
}

%$^3$EIHW -- Embedded Intelligence for Health Care and Wellbeing, University of Augsburg, Augsburg, Germany\\ removed according to Andreas

\markboth{Does the Definition of Difficulty Matter?}{Curriculum Learning}
\maketitle
\IEEEpeerreviewmaketitle

\begin{abstract}
\Ac{cl} describes a machine learning training strategy in which samples are gradually introduced into the training process based on their difficulty.
Despite a partially contradictory body of evidence in the literature, CL finds popularity in deep learning research due to its promise of leveraging human-inspired curricula to achieve higher model performance.
% \Ac{cl} is a popular topic in deep learning research as it promises higher model performance by leveraging human-inspired curricula, where samples are gradually introduced into the training process based on their difficulty.
Yet, the subjectivity and biases that follow any necessary definition of difficulty, especially for those found in orderings derived from models or training statistics, have rarely been investigated.
% Despite repeatedly reported benefits through \ac{cl}, some studies raise doubts about the universal suitability of \ac{cl}, leading to a partially contradicting body of evidence in the literature.
% Moreover, the majority of \ac{cl} literature has focused on one modality, namely vision.
To shed more light on the underlying unanswered questions, we conduct an extensive study on the robustness and similarity of the most common scoring functions for sample difficulty estimation, as well as their potential benefits in \ac{cl}, using the popular benchmark dataset CIFAR-10 and the acoustic scene classification task from the DCASE2020 challenge as representatives of computer vision and computer audition, respectively.
We report a strong dependence of scoring functions on the training setting, including randomness, which can partly be mitigated through ensemble scoring. %, leading to a higher agreement across scoring functions.
While we do not find a general advantage of \ac{cl} over uniform sampling, we observe that the ordering in which data is presented for \ac{cl}-based training plays an important role in model performance.
% , as the common easy-to-hard ordering leads to substantially higher performance than the reversed (anti-curriculum) ordering.
Furthermore, we find that the robustness of scoring functions across random seeds positively correlates with \ac{cl} performance. 
Finally, we uncover that models trained with different \ac{cl} strategies complement each other by boosting predictive power through late fusion, likely due to differences in the learnt concepts.
Alongside our findings, we release the \emph{aucurriculum} toolkit (\href{https://github.com/autrainer/aucurriculum}{https://github.com/autrainer/aucurriculum}), implementing sample difficulty and \ac{cl}-based training in a modular fashion.

\end{abstract}
%It has been a popular topic in deep learning research as it promises higher model performance by leveraging human-inspired curricula through the concept of sample difficulty.


\section{Introduction}
As for many other concepts in \ac{ml}, \acf{cl} owes parts of its appeal to its relatable inspiration from human learning: in the same way that children are first taught clear and distinguishable concepts before more difficult or nuanced ones~\cite{human-learn-curricula-bastea, human-teach-curricula-khahow}, \ac{ml} should also benefit from a curriculum structure, in which models are first confronted with easy examples before difficulty is steadily increased.
In the broader optimisation background of \acp{dnn}, this idea could be conceptualised as easier samples providing a smoother loss landscape, allowing models to quickly reach favourable parameter regions and finally converge therein~\cite{cl-bencur, cl-survey-wanasu}.
The most common implementations of \ac{cl} apply the underlying concept in the following way~\cite{score-ce-loss-hacont}: First, all samples of a given training dataset are sorted by difficulty.
Then, during training, the model is gradually exposed to more and more samples of increasing difficulty, beginning with an initial `easy' subset, with the full dataset incrementally introduced in later stages~\cite{cl-bencur, cl-survey-wanasu}.

Despite the intuitive motivation behind \ac{cl}, defining or determining a measure of \ac{sd} to create such a difficulty ordering remains a key open challenge~\cite{cl-survey-wanasu, cl-survey-sovcur, cl-survey-liuare}.
This aspect is so crucial that any investigations into the potential benefits of \ac{cl} depend on a well-founded quantification of \ac{sd}.
Without at least a robust contextual sense of difficulty, the core assumption for the concept of \ac{cl} vanishes.
Additionally, a universally applicable human-inspired difficulty understanding is still missing despite efforts in this direction~\cite{diff-mayhow,wang2020survey}.
Moreover, a straightforward transfer of human difficulty to \ac{cl} is debatable due to significant differences in learning between humans and machines~\cite{song2024inferring}.
Instead, many model-based approaches to \ac{sd} estimation have been suggested in recent years. %, all of which essentially rely on the model training dynamics of a specific learning setting.
% Although impacts of the chosen model architecture, training parameters, and even randomness on the difficulty estimation can be expected, no study -- to this day -- has been concerned with a thorough and quantitative evaluation of these effects.
Yet, despite the fact that these approximations depend on training setups (\eg model architecture, hyperparameters, or even random effects due to initialisation), there is no comprehensive study quantifying their impact.
This is of particular importance given the recent trend of using \ac{sd} to garner insights into the structure of datasets~\cite{diff-medtri}.
We aim to overcome this limitation by exploring various configuration settings that shed more insight into the impact of these different factors.

Furthermore, despite the fact that evidence of the benefits of \ac{cl} has been reported ``en masse''~\cite{score-ce-loss-hacont, cl-ensemble-kesdev, cl-related-lotcur, mallol2020curriculum}, the methodology has not found its way into the standard training procedures.
It is not clear whether this is due to the additional computational overhead and implementation complexity or an overestimation of promised benefits due to the factors mentioned above. 
% Whether the reasons for this lie in additional overhead in computational and implementation complexity or an overestimation of the promised benefits, for instance, due to model selection biases in reported studies or limited generalisability across datasets, remains unclear.
A promising attempt to analyse contexts in which \ac{cl} proves beneficial is provided by~\cite{misc-when-do-curricula-work}.
The study compares various \acp{sf} for the estimation of \ac{sd}, \acp{pf} for the scheduling of samples during training, and different difficulty orderings (\eg easy-to-hard, hard-to-easy, and random), providing evidence that the benefits of \ac{cl} are mainly limited to shorter training time or mitigating label noise.
However, the authors focus on a coarse statistical view of the problem, leaving out a more granular analysis of the individual aspects of \acp{sf} and their interplay with \ac{cl} settings.
Moreover, with an isolated performance analysis of the trained models, it remains an open question as to how training dynamics influence the concepts learnt by the final model states through the use of \ac{cl} strategies -- a question we additionally explore in this work.

In our study, we also try to overcome a common limitation in the selection of datasets for large-scale \ac{cl} studies (and fundamental \ac{ml} studies in general), which heavily focus on standard \ac{cv} benchmark datasets~\cite{cl-survey-sovcur, dl-sharpness-lihvis}.
To sustain comparability to existing literature, we perform our experiments on one of the most popular \ac{cv} datasets, \mbox{CIFAR-10}. We further extend them to a popular \ac{ca} task of \ac{asc} to potentially unravel modality-specific differences, as suggested in prior studies~\cite{triantafyllopoulos2021role, milling2024bringing}.

In summary, we aim to present a thorough analysis of properties and similarities for an extensive subset of popular \acp{sf} for \ac{sd} estimation.
We further evaluate the interplay of difficulty orderings with \ac{cl} and analyse synergies between model states trained with different \ac{cl} strategies. 
In the process, we try to bring new insights to the advantages, the limitations, and the general workings of curriculum-based learning paradigms by, at least partially, answering the following questions:
\begin{itemize}
    % (old version)
    %\item How robust or susceptible to variation are different \acp{sf} concerning different choices of model architectures, training hyperparameters and random seeds, and what are their limitations in a practical setting?
    \item How sensitive are different \acp{sf} to varying model architectures, training settings, as well as random seeds, and what are their limitations in a practical setting?
    % Alternatives for robustness: sensitivity, stability, consistency, variability
    \item To what extent do different \acp{sf} share a similar notion of \ac{sd}?
    % (old version)
    %\item Are more robust \acp{sf} advantageous for \ac{cl}?
    \item Are less training setting-sensitive \acp{sf} advantageous for achieving higher performance?
    \item How do different \ac{cl} strategies impact model performance?
    %\item Can models trained with different \ac{cl} strategies be combined for improved predictive performance?
    \item Do models trained with varying \ac{cl} strategies learn different concepts that can be combined to improve their predictive performance?
\end{itemize}
The remainder of this work is structured as follows: \Cref{sec:related} presents a thorough overview of the most common \acp{sf} and \ac{cl} studies.
\Cref{sec:meth} introduces the datasets and model architectures that are investigated in this work, as well as the selection and modifications applied to \acp{sf} and \ac{cl} strategies.
The results of our study are presented and discussed in \Cref{sec:exp} and finally summarised in \Cref{sec:conc}.

\section{Related Work}
\label{sec:related}

\subsection{Scoring Functions}
Related literature offers several approaches to estimate the difficulty of each sample in the training set, primarily discussed in the context of \ac{cl}.
% Conceptually, there are several different categories for the quantification of difficulty notions.
These can be conceptually split into two different categories: human- and model-based difficulty.
Human-based difficulty definitions fall under two main paradigms:
\begin{itemize}[align=left]
    \item[\emph{Human priors}] One straightforward strategy is to base the estimation on human priors of difficulty understanding, which are easy to define and implement.
    Such attempts may include the length of a sentence for \ac{nlp} tasks~\cite{spitkovsky2010from, platanios2019competence, tay2019simple}, or the \ac{snr} in a signal~\cite{braun2017acurriculum,ranjan2018curriculum} for \ac{ca}~\cite{cl-survey-wanasu}.
    Despite their intuitive definitions, these difficulty estimations are often limited by the tendency to disregard the complexity of the data structures, inter-class relationships, and the fact that human- and model-based difficulty perceptions may differ.
    \item[\emph{Labelling efforts}] A more direct approach incorporating a human understanding of difficulty into the \ac{sd} estimation involves considering the human labelling behaviour for individual samples.
    This approach is particularly advantageous for datasets lacking ground truth labels due to the subjectivity of the task, such as in emotion recognition~\cite{schuller2018speech}.
    Here, gold standards are often formed for each sample based on a summary of the varying ratings assigned by different human annotators~\cite{ringeval2019avec}.
    The inter-annotator agreement can then be interpreted as a measure of difficulty, with harder samples expected to result in greater variation among ratings~\cite{cl-related-lotcur}.
    Alternatively, humans can perform an explicit (direct or indirect) rating of the \ac{sd}.
    As reported in~\cite{diff-mayhow}, the \ac{mvt} necessary for a human to categorise an image can be interpreted as a good proxy for difficulty.
    In regression tasks, the targets -- in terms of their distance to the mean of the value range -- can be considered an estimate for difficulty \cite{mallol2020curriculum}.
\end{itemize}

The concrete examples of \acp{sf}, however, that we primarily focus on in this work are model-based.
These can be employed in the context of \ac{dnn}-based classification tasks on any data type without requiring additional human annotations beyond the target labels.
A plethora of these approaches exist in the literature, all of which generally share the idea of interpreting \ac{sd} through the lens of trained \ac{ml} models~\cite{cl-survey-liuare, cl-survey-wanasu, cl-survey-rl-narcur}.
While we refer to \Cref{sec:meth_scoring_functions} for a detailed definition of the \acp{sf} relevant to this work, some of the implemented concepts for sample-wise difficulty scores include: the expected performance on the sample if withheld from training, the epoch at which the sample is learnt, the consistency with which the sample is correctly classified during training, the value of the loss function on the sample after training, the \ac{dnn} layer at which the classification of the sample's intermediate representation aligns with the final model's prediction, and the separability of the sample's features obtained from a pre-trained model.

Despite the abundance of approaches (and to some degree arbitrary choices like network architectures and training parameters), little effort has been spent to analyse the robustness and similarity of different \acp{sf}: Wu \etal \cite{misc-when-do-curricula-work} investigate some of the \acp{sf} further analysed in this work and discover that difficulty estimations vary more across architecture families -- like \acp{fcnn} and \acp{cnn} -- than within. 
Beyond that, they observe a moderate to high correlation of the \ac{sd} across similar architectures for the chosen \acp{sf}.
Further, Mayo \etal \cite{diff-mayhow} explore the likeness of the \ac{mvt} based on human image recognition capabilities and a subset of model-based \acp{sf}. They discover a reasonable similarity across methods, yet with some limitations indicating differences in the understanding of \ac{sd} between humans and machines.

\subsection{Curriculum Learning}
Experimental studies surrounding the potential benefits of \ac{cl}-based learning paradigms are naturally limited through computational resource demands to specific contexts and settings, with the most extensive examples being restricted to \ac{cv} datasets.
Consequentially, contradicting evidence is reported in some cases across different works, leaving the question of the advantages of \ac{cl} open.

For instance, Hacohen and Weinshall~\cite{score-ce-loss-hacont} report, compared to the baseline, a higher performance for models trained with a curriculum considering two \acp{sf}: one is based on pre-trained feature extraction, while the other one leverages the loss values of the final model state.
However, a form of \ac{spl}, \ie a method in which a model determines and dynamically adjusts the \ac{sd} during training, performed worse than the baseline training setup.
Amongst the chosen \ac{cv} datasets, the advantages of \ac{cl} were deemed higher for more difficult datasets.
Beyond, the results indicate that the most impact of \ac{cl} happens at the earlier stages of the training.

On the other hand, the likely most extensive \ac{cv}-based study with respect to the number of trained models~\cite{misc-when-do-curricula-work} is purely based on a set of model-based \acp{sf}.
It considers ensembles to achieve higher stability in the difficulty orderings.
It agrees with the conclusion that \ac{cl} shows faster learning speed at the beginning of the training but cannot find evidence for a significantly higher performance enabled through \ac{cl} in any setting with a fair comparison to the baseline.
As a random \ac{sd} ordering is reported to yield similar performance benefits as any order obtained from the applied \acp{sf}, it is hypothesised that the advantages of \ac{cl} can be attributed primarily to the dynamic dataset size during training.
In particular, \acp{pf} tend to perform better when they quickly incorporate more difficult samples and saturate on the full dataset.

Beyond the realm of \ac{cv}, Lotfian \etal \cite{cl-related-lotcur} perform \ac{cl} experiments for \ac{ser}.
The study considers both model-based \acp{sf} obtained from the models' loss functions as well as human-based \acp{sf} derived from the inter-rater agreement.
The authors report marginally higher performance for the model-based ordering and noticeably higher performance for the rater-based ordering, compared to the baseline.
Contrary to Wu \etal \cite{misc-when-do-curricula-work}, they do not find any comparable benefits from random orderings.
A study on \ac{cl} for \ac{asr} was performed by Karakasidis \etal \cite{cl-related-karcom}, which considers difficulty-orderings based on the utterance length, a model-based \ac{sf}, and \ac{spl}.
The length of the utterance, which aims to incorporate human priors into the \ac{sd} estimation, achieved the lowest performance overall.
Contrary to Hacohen and Weinshall~\cite{score-ce-loss-hacont}, the best results are obtained with \ac{spl}, yet \ac{cl}-based approaches still performed on par.
It is further reported that the benefits of employing \acp{pf} are particularly eminent in the later stages of the training, contradicting both~\cite{score-ce-loss-hacont} and~\cite{misc-when-do-curricula-work}.

The non-uniform conclusions reported in this section may originate in the vastly different settings in which the studies are performed. Many variations can be observed in terms of \acp{sf}, \acp{pf}, datasets, model architectures, as well as the training routines and how baseline comparisons are defined.
The studies further fail to analyse the characteristics of the suggested \acp{sf} and the corresponding importance for \ac{cl} experiments, which we plan to address in the following.

\section{Methodology}
\label{sec:meth}
To answer the research questions posed in the introduction, we investigate a total of six \aclp{sf}, three \acl{sd} orderings, and four \aclp{pf} over two datasets and five \ac{dnn} models.
In the following subsections, we elaborate on each of these aspects in detail.

\subsection{Datasets}
\label{sec:met_datasets}
Our experiments are performed on two 10-class classification datasets from the domains of \ac{ca} and \ac{cv}, respectively.
Both datasets vary in size, comprise samples with different representation sizes, and have different baseline accuracies.

\textbf{CIFAR-10}: The subset of the tiny image recognition dataset referred to as CIFAR-10~\cite{krizhevsky2009learning} is one of the most popular benchmark datasets for understanding \ac{dl} training in general and for \ac{cv} in particular~\cite{glorot2010understanding, zhang2017mixup}.
It comprises 60\,000 images, divided into 50\,000 samples for training and 10\,000 for testing, with both subsets being fully balanced across ten classes, which can coarsely be categorised as animals and vehicles.
The images are characterised by a rather small representation size of 32$\times$32 pixels with three colour channels.
Modern \sota approaches are able to achieve very high performance on the dataset~\cite{model-efficientnet, model-efficientnet-v2, model-vit, model-conv2next}.
We upsample all images using bilinear interpolation to balance accuracy and computational costs. 
This results in a representation size of $ 64 \times 64 \times 3$.
% Our baseline architectures without excessive \mbox{(hyperparameter-)optimisation} of the training methodology or pre-training achieve well beyond 80\,\% accuracy.
It is important to note that we employ a train-test split without an explicit validation subset, as the main focus of our work is on understanding \ac{sd} on the full training dataset rather than achieving \sota performance.
This approach is applied consistently across both datasets, as well as the baseline and \ac{cl}-based training experiments.

\textbf{DCASE2020}: For the \ac{ca} dataset, we choose task 1a of the DCASE2020 challenge~\cite{Heittola2020}.
It comprises 13\,962 training and 2968 testing audio samples of 10\,s length across many different cities, which are recorded with different real and simulated devices.
Both the training and testing subsets are nearly class-balanced.
The labels indicate the acoustic scene, representing the type of location where the audio samples were recorded.
Examples are public transport vehicles, open spaces, and indoor environments.
In our experiments, raw audio samples are transformed into log Mel-spectrograms with 64 Mel bins, a window size of 512\,ms, and a hop size of 64\,ms, following the spectrogram extraction process outlined in~\cite{model-cnn-10-14}.
This results in a representation size of 64$\times$1001$\times$1, where each element corresponds to the magnitude of the signal's energy in a specific Mel-frequency bin and time frame.
% Compared to CIFAR-10, the dataset can be considered more difficult, as our baseline experiments without pre-training only achieve accuracies close to 60\,\%.

\subsection{Network architectures}
\label{sec:meth_networks}
Given the image-like nature of both the CIFAR-10 samples and the log-Mel spectrograms in the DCASE2020 datasets, we decide to employ \ac{cnn}-based architectures, which are well-established in the literature of the respective tasks~\cite{dl-survey-donasu, dl-audio-survey-dinaco}.

\textbf{ResNets}: \acp{resnet}~\cite{model-resnet} are among the most impactful \ac{dnn} architectures in \ac{dl}, playing a significant role up to this day.
The introduction of residual (or skip) connections allowed for developing very deep \ac{cnn} architectures.
In our experiments, we specifically employ the \ac{resnet}50 architecture for the CIFAR-10 dataset.

\textbf{EfficientNets}: The EfficientNet family~\cite{model-efficientnet} is inspired by residual connections of the \ac{resnet} architecture and further employs mobile inverted bottlenecks~\cite{model-mobilenet-v2} as well as squeeze-and-excitation blocks~\cite{model-squeeze-and-excitation-net}.
The base model, EfficientNet-B0, is optimised to balance task accuracy and computational complexity, while larger versions are derived via compound scaling of the architecture's depth, width, and resolution.
In our experiments, we utilise EfficientNet-B0 for both the CIFAR-10 and DCASE2020 datasets, and the larger EfficientNet-B4 only for the CIFAR-10 dataset.

\textbf{PANNs}: The CNN10 and CNN14 architectures were introduced as part of the \ac{pann}~\cite{model-cnn-10-14} family for spectrogram-based \ac{ca} tasks.
These models follow a more traditional \ac{cnn} design, comprising regular convolutional and batch normalisation layers without skip connections, and are inspired by the VGG architecture~\cite{model-vgg}.
CNN14 can be considered a scaled-up version of the CNN10 architecture.
Both models are only utilised for the DCASE2020 dataset.

Overall, we consider two types of initialisation for all presented networks: randomly initialised and pre-trained on the large datasets ImageNet~\cite{russakovsky2015imagenet} for the \ac{resnet} and EfficientNet architectures, and AudioSet~\cite{ds-audioset} for the \ac{pann} architectures.

\subsection{Scoring Functions}
\label{sec:meth_scoring_functions}
\textbf{Consistency Score (C-score)}: The concept best-suited to a model-based \ac{sd} quantification in a classification setting is the \emph{C-score}, as introduced in~\cite{score-c-score}.
Its core idea is to measure how consistently a sample is classified correctly across different training subsets of varying sizes, with the sample in question excluded from the training process.
Samples that can consistently be classified with a small amount of training data likely require less data complexity, are more representative of their class, and are, therefore, easier to classify.
However, obtaining a robust version of the C-score is limited by the computational costs required to train at least one model for each subset size and sample excluded from training.
By definition, a perfect calculation of the C-score would require several evaluations across all possible subsets of the training data.
Since this would lead to a prohibitively large amount of experiments, the C-score can only be approximated.
Jiang \etal \cite{score-c-score} draw 2000 subsets for each subset ratio $s \in \{ 10\%, \dots, 90\% \}$, creating publicly available proxies of the C-score\footnote{\url{https://pluskid.github.io/structural-regularity/}} for the CIFAR-10 dataset based on 18\,000 models.
Given the high computational costs associated with the calculation of the C-score, we limit our analysis to the publicly available C-scores for CIFAR-10 and do not determine them for the \ac{asc} task.

\textbf{Cross-Validation Loss (CVLoss)}: As a computationally cheaper yet still expensive alternative to the C-score, we draw inspiration from the \ac{cvst} \ac{sf}~\cite{misc-when-do-curricula-work} and a loss-based C-score variant~\cite{cl-ensemble-kesdev}, defining the \ac{cvloss} as the average loss of a sample when it is held out in a randomly partitioned $k$-fold cross-validation setting.
The loss-based version allows for a more fine-grained \ac{sd} estimation compared to the accuracy-based C-score but does not account for the effects of differently sized training subsets. 
In our experiments, we use $k=3$ partitions, allowing us to obtain one score for each sample from the training of three models.
This approach achieves a reasonable trade-off between computational cost and the model's expected performance given the size of the training subset.

\textbf{Cumulative Accuracy (CumAcc) and First Iteration (FIT)}: More computationally lightweight examples of \acp{sf} include \ac{cumacc}~\cite{score-c-score} and \ac{fit}~\cite{score-first-iteration, misc-when-do-curricula-work}.
Both approaches rely on the learning statistics of each sample during the training process.
They allow for the \ac{sd} estimation of all training samples from a single model training.
\ac{cumacc} is calculated as the sum of correct classifications over the total number of training epochs.
In contrast, \ac{fit} is defined as the ratio of the first correct epoch, in which a sample is correctly classified and remains correct thereafter, and the total number of epochs.
In this context, easier samples are expected to be learnt earlier and classified correctly with greater consistency.

\textbf{Cross-Entropy Loss (CELoss)}: Another \ac{sf} can be inferred from the loss of each training sample in a single trained model state.
Examples with a lower loss after training are interpreted as easier, as the model can fit them more effectively.
Although this approach was originally introduced as \textit{bootstrapping} in~\cite{score-ce-loss-hacont}, we refer to this \ac{sf} as \ac{celoss} for clarity, particularly in the context of the classification tasks investigated in this work.

\textbf{Transfer Teacher (TT)}: Similarly, \ac{tt}~\cite{score-transfer-teacher} estimates \ac{sd} by analysing the classification boundaries of the samples after training.
In contrast to \ac{celoss} and previous \acp{sf}, however, \ac{tt} does not evaluate \ac{dl} models trained on the target task.
Instead, \ac{tt} extracts features from the penultimate layer of a model pre-trained on a larger dataset,  % -- often \ac{dl} models trained on large foundation datasets like ImageNet~\cite{russakovsky2015imagenet} -- 
which are then classified via a \ac{svc}.
The margins between the samples and the classification boundaries indicate the estimated \ac{sd}.
The underlying idea is that the \ac{svc} should place easy samples far from the boundaries.

\textbf{Prediction Depth (PD)}: Another combination of \ac{dnn}-based representations and traditional machine learning algorithms for classification is the \ac{pd} introduced in~\cite{score-prediction-depth}. 
\ac{pd} looks at the representations extracted from the final state of a trained model at different layers, which supposedly represent different views of the same data sample in the form of low- to high-level features.
The representations extracted from the layers are then classified by \ac{knn} probes with $k=30$.
Baldock \etal \cite{score-prediction-depth} place probes both at the input of a model and after applying the softmax function to the output.
For \ac{resnet} architectures, probes are positioned after the normalisation layer in the stem and following the skip connection in each residual block.
For VGG architectures, probes are placed after each convolutional layer.
We extend the \ac{resnet} probe placement strategy to the EfficientNet family by inserting probes after each block.
As both CNN10 and CNN14 are inspired by the VGG architecture, we place probes after each normalisation layer and the first linear layer.
This results in 19 probes for ResNet50, 20 for EfficientNet-B0, 36 for EfficientNet-B4, 12 for CNN10, and 16 for CNN14.
%For \ac{resnet} models, probes are placed at the model input, after the normalisation in the stem, after each residual block, and following the final output of the network.
%As no general rule for the probe placement is provided, we utilise this strategy for all \ac{cnn} architectures.
The \ac{pd} is defined as the depth of the first layer in which the \gls{knn} probe's prediction matches the model's final prediction and aligns for all subsequent probes in deeper layers.
Easier samples are expected to be separable at lower layers, while harder samples should require higher-level features for classification.
While Baldock \etal \cite{score-prediction-depth} do not assign a \ac{sd} score if the prediction of the final probe differs from the network's prediction, we assign the maximum depth to obtain a difficulty measure for the complete training subset.
To ensure that the \ac{knn} probes are computationally feasible, we limit the representation size at each layer to 8192.
If any representation exceeds this limit, \ac{gap} is applied to the spatial dimensions of the input, while the number of channels is preserved.
Beyond, \ac{gap} is applied to match the spatial dimensions of the input, allowing for an equal contribution of time and frequency information in the case of our \ac{asc} task.
However, the use of \ac{gap} may lead to the loss of critical features, potentially challenging the \ac{sf}'s ability to accurately estimate the \ac{sd}.

\subsection{Difficulty Ordering and Ensemble Scoring}
\label{sec:meth_ensemble}
Any of the \acp{sf} introduced in \Cref{sec:meth_scoring_functions} can assign a difficulty score to each individual sample in the training set, however, these scores are generally not normalised.
For the purpose of \ac{cl}, as we apply it, the exact difficulty score is not of particular importance.
Instead, we solely rely on a difficulty-based ordering derived from the \ac{sd} scores provided by the respective \ac{sf}.
In cases where multiple samples share the same exact difficulty score, they are sorted according to the original dataset ordering to avoid introducing randomness into the process.
Ensembles across multiple \acp{sf} of the same type can be constructed by averaging the \ac{sd} scores from each \ac{sf} prior to creating the final difficulty ordering~\cite{cl-ensemble-kesdev}.

\subsection{Pacing Functions}
\label{sec:meth_pacing_functions}
A \ac{cl} experiment, as defined in~\cite{misc-when-do-curricula-work} and implemented in this work, considers a \ac{sd} ordering and starts a gradient-based optimisation on the first fraction $b$ of the training set according to said ordering.
For comparison, the experiment may also use a reversed \ac{sd} ordering (referred to as \ac{acl}) or a random ordering (referred to as \ac{rcl}).
In the case of \ac{cl} and $b=0.2$, the first fraction corresponds to the 20\,\% easiest examples of the training set.
Throughout the training, the size of the dataset is monotonically increased according to the provided \ac{sd} ordering until a fraction $a$ of the total training iterations is reached, after which the full training set is utilised.
For the phase between the training on the initial subset $b$ and the full dataset, we employ several functions that monotonically increase the size of the subset.
All \acp{pf} share the same boundary conditions, defined by $b$ and $a$, for the initial dataset size and the iteration at which the full dataset is employed.
However, some \acp{pf} add samples more slowly at the beginning and faster towards the end, while others start by adding samples quickly before slowing down as they approach $a$.
Ordered from the fastest to the slowest dataset size increase at the beginning of the training, we utilise some of the \acp{pf} defined according to Wu \etal \cite{misc-when-do-curricula-work}: \textit{logarithmic} (log), \textit{root}, \textit{linear}, and \textit{exponential} (exp).

\section{Experiments and Discussion}
\label{sec:exp}
In the following, our experiments and corresponding discussions are structured sequentially, first focusing on the behaviours of \acp{sf}, then analysing the performance of curriculum-based training settings.
All experiments are conducted using Python 3.10 and PyTorch 2.1.0~\cite{misc-pytorch-framework}.
For any hyperparameter not explicitly modified, the default values provided by PyTorch are used.
Additionally, we release the \emph{aucurriculum}\footnote{\url{https://github.com/autrainer/aucurriculum}} toolkit, built on top of \emph{autrainer}\footnote{\url{https://github.com/autrainer/autrainer}}, which implements all \acp{sf} and \acp{pf} explored in this work.
This package allows for obtaining \ac{sd} scores from arbitrary classification datasets and supports curriculum-based training.
The code to reproduce our experiments is publicly available on GitHub\footnote{\url{https://github.com/ramppdev/sample-difficulty-curriculum-learning}}.
Training is performed across a variety of GPU architectures, including NVIDIA GeForce GTX TITAN X, GeForce RTX 3090, and A40 GPUs.
Given the diverse nature of the utilised GPUs, we omit an analysis of training time comparisons between the standard training baseline and curriculum-based experiments, as this anyway depends on hardware specifications.

\subsection{Baselines}
\label{sec:exp_baselines}
To identify the set of hyperparameters best suited to each model and dataset and establish a baseline to compare with, we run a set of preliminary experiments iterating over a small set of hyperparameters (\cf \cref{tab:grid_search}).
In all cases, models are trained for the full 50 epochs, and the final model selection is based on the best-performing model state on the respective validation set.
Overall, we train 54 models per dataset, and the best-performing training configuration for each model and initialisation is reported in the appendix (\cf \Cref{tab:performance_baselines}).
% The best configurations achieve a performance of 0.949 for CIFAR-10 using a pre-trained ResNet50 with the SAM optimiser and a learning rate of 0.001, and 0.678 for DCASE2020 employing a pre-trained CNN14 with the Adam optimiser and a learning rate of 0.0001.
Even though the best performances for CIFAR-10 and DCASE2020 are obtained with pre-trained versions of ResNet50 and CNN14, respectively, 
we select the best-performing EfficientNet-B0 configuration for both datasets as a baseline reference for further analysis.
This choice is motivated by the model being the only architecture employed for both datasets and its comparably low parameter count, allowing for a more efficient training.
Training the model from scratch further reduces the potential impact of pre-training data. 
The reference baseline performances are thus 0.839 for CIFAR-10 and 0.583 for DCASE2020, achieved using the Adam optimiser with a learning rate of 0.001 and the SAM optimiser with a learning rate of 0.01, respectively.
Nonetheless, we consider all models to investigate the impact of varying training settings on \acp{sf} in \ref{sec:exp_scoring}.

\subsection{Scoring Functions}
\label{sec:exp_scoring}
\subsubsection{Impact of Training Setting}
\label{sec:exp_scoring_robustness}
Our first analysis builds on top of the previous baseline experiments and the obtained model states.
We focus on how much the choice of different training settings affects the final difficulty ordering provided by certain \ac{sf}.

We first create different \ac{sd} orderings for the \acp{sf} presented in \Cref{sec:meth_scoring_functions} by independently varying the random seed, the model architecture (including initialisation), and the optimiser-learning rate combination, respectively.
Each variation is based on the best-performing EfficientNet-B0 reference configuration, with only one setting -- seed, model, or optimiser-learning rate combination -- varied at a time for comparison.
Overall, we choose six different variations per training setting and obtain one \ac{sd} ordering per \ac{sf} and variation.
The variations are summarised in the appendix (\cf \Cref{tab:scoring_function_variations}) and follow the respective choices from \Cref{tab:grid_search}, but we add five new random seeds to obtain variations \wrt randomness.
Additionally, we limit our analysis to the two best-performing learning rate combinations for each optimiser, as some combinations did not converge during training.
Note that the random seed controls both the model initialisation and the ordering of training data during epochs.
This guarantees that the results are reproducible, and we further aim to minimise the influence of randomness on model or optimiser variations.

% 53 models: 5 seeds (for all training-based SFs) + CVLoss (6 seeds + 5 models + 5 optimisers = 16) * 3 (k=3) = 53 total, one of model and optim can be reused for CVLoss

While most \ac{sd} orderings can be obtained straight from the baseline experiments performed in \Cref{sec:exp_baselines}, an additional 53 models need to be trained per dataset due to the added consideration of random seeds and the requirements of the \ac{cvloss} \ac{sf} calculation.
To reduce the computational costs of the \ac{cvloss} \ac{sf}, we reduce the training time to 25 epochs.
Moreover, \ac{tt} is only calculated over the pre-trained versions of the three model architectures, as the \ac{sf} requires a pre-trained model by design, and the \ac{svc} fitting on the target task does not allow for varying random seeds or optimisation routines.
Whenever the selection of a specific model state is required to calculate a \ac{sf}, we choose the best-performing state analogous to \Cref{sec:exp_baselines}.

\Cref{tab:scoring_function_robustness} summarizes the \ac{sf} `robustness' per training setting for CIFAR-10 and the DCASE2020, respectively.
We compute the mean and standard deviation of pairwise Spearman rank correlations of the difficulty orderings obtained for each independently varied training setting.
From the six variations within each training setting (seed, model, and optimiser-learning rate combination), we obtain 15 unique pairwise correlations (excluding self-correlations).

The mean pairwise correlation is, in most cases, moderate ($\geq 0.4$) or strong ($\geq 0.6$).
This is true for both datasets, a fact that indicates moderate to high agreement for all \acp{sf} when changing one training setting and, thus, a consistent underlying notion of \ac{sd} for each \ac{sf}.

With respect to the type of training setting, varying the random seed results in overall higher agreement, with a change in the model architecture exhibiting the lowest agreement.
This shows how different models -- with different inductive biases -- encapsulate a different notion of \ac{sd}, with important training settings such as the optimiser and learning rate adding further variability.
Nevertheless, even changing the random seed alone -- and keeping all other things equal -- results in a low to moderate amount of disagreement.
As the random seed controls both the model initialisation and the order in which examples are presented during training, it is evident that these two aspects of training play a non-negligible role when discussing \ac{sd}.

In terms of comparing the two datasets, we observe that they exhibit differences in both directions, with CIFAR-10 showing higher correlations than DCASE2020 in some constellations and lower in others.
This shows how the dataset plays an additional role for the robustness of a \ac{sf} across different training settings and is an extra source of variability.

In summary, we conclude that an \ac{sd} ordering is context-dependent, as it is influenced by the choice of model, training hyperparameters, and even random seed selection.
The corollary is that the model-based curricula we investigate later can be seen as `global' curricula in only a relative sense, as each training setting results in a different \ac{sd} ordering.

\begin{table}[t]
\centering
  \caption{
  Training setting robustness in terms of agreement of difficulty rankings obtained with varying training settings. 
  The respective training setting of each column is varied with respect to the reference configuration.
  We then report the mean and standard deviation of Spearman correlation coefficients ($\rho$) computed between unique pairs of training setting variations.
  }
  \label{tab:scoring_function_robustness}
  \begin{tabular}{lrrr}
    \toprule
    \textbf{SF} & \textbf{Seed} & \textbf{Model} & \textbf{Optim + LR}\\
    \midrule
    \multicolumn{4}{l}{\textbf{CIFAR-10}} \\
    \midrule
    CELoss & .507 $\pm$ .026 & .428 $\pm$ .043 & .483 $\pm$ .055\\
    CVLoss & .676 $\pm$ .007 & .629 $\pm$ .045 & .689 $\pm$ .022\\
    CumAcc & .760 $\pm$ .008 & .557 $\pm$ .101 & .752 $\pm$ .019\\
    FIT & .586 $\pm$ .033 & .416 $\pm$ .076 & .623 $\pm$ .019\\
    PD & .790 $\pm$ .012 & .653 $\pm$ .076 & .799 $\pm$ .032\\
    TT & -- & .648 $\pm$ .025 & --\\
    \midrule
    \multicolumn{4}{l}{\textbf{DCASE2020}} \\
    \midrule
    CELoss & .410 $\pm$ .060 & .415 $\pm$ .115 & .369 $\pm$ .041 \\
    CVLoss & .591 $\pm$ .018 & .579 $\pm$ .044 & .556 $\pm$ .046 \\
    CumAcc & .821 $\pm$ .012 & .590 $\pm$ .099 & .758 $\pm$ .048 \\
    FIT & .604 $\pm$ .020 & .475 $\pm$ .084 & .513 $\pm$ .052 \\
    PD & .748 $\pm$ .023 & .694 $\pm$ .046 & .683 $\pm$ .068 \\
    TT & -- & .523 $\pm$ .191 & --\\
    \bottomrule
  \end{tabular}
\end{table}

\subsubsection{Robustness of Ensembles}
\label{sec:exp_scoring_ensemble}
The strong influence of randomness across all \acp{sf} observed in \Cref{sec:exp_scoring_robustness} poses some concerns towards the legitimacy of model-based \ac{sd} prediction.
The impacts of variations in model architecture or optimisation routines could be attributed to inherent biases, resulting in differing interpretations of \ac{sd}.
However, the role of randomness through varying seeds is harder to justify and, to some extent, contradicts the concept of \ac{cl} as it prevents an objective difficulty ordering.

A possible mitigation strategy for this limitation might lie in the use of ensemble scoring functions, following the experiments of~\cite{misc-when-do-curricula-work} and~\cite{cl-ensemble-kesdev}.
The underlying hypothesis of this approach is that the effects of random variations can be counteracted by averaging over several difficulty predictions obtained from different random seeds for each sample, thus unravelling a more robust and less biased \ac{sd} estimation.
In order to investigate this hypothesis, we first train a total of nine new models per dataset by varying the random seed of the EfficientNet-B0 reference configuration.
Per dataset and \ac{sf}, we obtain 15 difficulty estimations, the settings of which only vary in the underlying random seed. We omit \ac{cvloss} due to computational costs and \ac{tt} due to a lack of dependence on randomness from the discussion.
From the 15 difficulty estimations, we form three ensemble difficulty orderings for each of the varying ensemble sizes \{1,2,3,4,5\}, as described in \Cref{sec:meth_ensemble}, while ensuring no overlap in the underlying \acp{sf} across the ensembles.

\begin{figure}[t]
    \centering
    \includegraphics[width=\columnwidth]{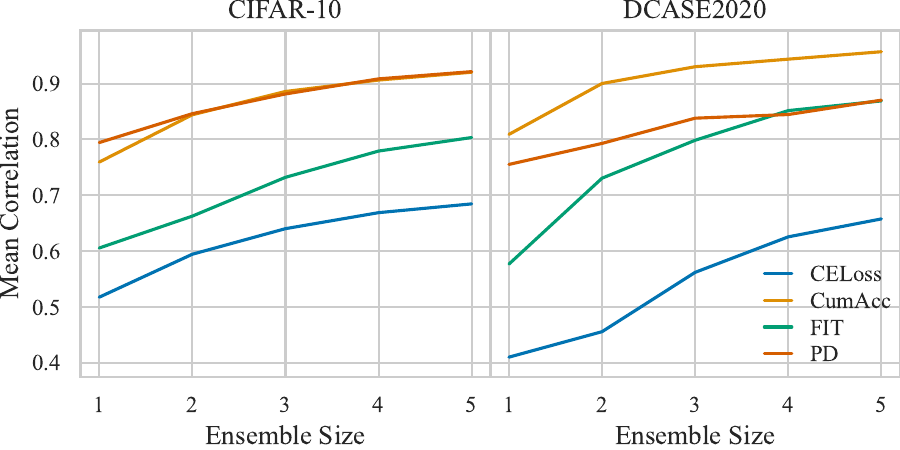}
    \caption{
    Correlation of \acp{sf} with increased ensemble sizes for the datasets CIFAR-10 and DCASE2020.
    The ensemble size encapsulates how many individual orderings -- obtained from different random seeds -- are considered to build one ensemble.
    For each \ac{sf}, we report pairwise Spearman correlations across three ensemble orderings with the same ensemble size.
    }
    \label{fig:seed_aggregation_correlation}
\end{figure}

\Cref{fig:seed_aggregation_correlation} illustrates the average pairwise correlation across the three considered ensemble difficulty orderings for each ensemble size.
It is apparent that the pairwise correlation consistently increases with larger ensemble sizes across datasets and \acp{sf}.
\Ac{sd} orders obtained from the same experimental settings but different random seeds thus agree more with each other if the orders are built via an ensemble over multiple random seeds.
%The only outlier to this trend is \ac{pd}, for which extraordinarily high correlations are noted to begin with, and which are further discussed in the next section.
This result is in line with our hypothesis, clearly indicating that difficulty orderings become more robust towards randomness with increasing ensemble sizes.

\subsubsection{A Deeper Look into Difficulty Distributions}
\label{sec:exp_scoring_distribution}
A misleading conclusion that might be drawn from \Cref{sec:exp_scoring_robustness} and \Cref{fig:seed_aggregation_correlation} would be that higher training setting robustness would automatically be a sign for a better \ac{sf}. 
However, there are some methodological decisions limiting this interpretation: A comparison of difficulty orderings assumes unique and distinct difficulty values for each sample, which cannot be guaranteed through the \acp{sf}. 
% The actual difficulty estimations for each sample are not unique across samples. 
% The very high stability of \ac{pd} reported in \Cref{sec:exp_scoring_robustness} across randomness and the choice of optimisation routine might at first glance suggest that this \ac{sf} proves to be particularly robust and thus superior to other functions.
% This conclusion, however, is originating from another effect: while, in principle, a comparison of difficulty orderings assumes individual and distinct difficulty values for each sample, the actual difficulty estimations for each sample are, in fact, not unique across samples. 

In order to still provide a necessary, ranked difficulty ordering for \ac{cl} experiments, we introduced in \Cref{sec:meth_scoring_functions} the decision to rank samples with an identical difficulty estimation by the index provided by the original dataset.
This decision has its reason mainly in the exclusion of further randomness effects from \ac{cl} experiments.
It also ensures the similarity of two difficulty estimations, which assign the same difficulty values to various samples.
However, this decision has implications for the comparison of orderings with coarse granularity, as it assigns the same difficulty value to many samples. 

An overview of the granularity of the different \acp{sf} in the single-score as well as the ensemble context is given by \Cref{tab:difficulty_distributions}.
Naturally, the \acp{sf} \ac{fit}, \ac{cumacc}, and \ac{pd} show the coarsest granularity, as their discrete nature only allows for a limited number of difficulty predictions -- at least in the single-score context. 
% However, while \ac{cumacc} and \ac{fit} can `benefit' in terms of expressiveness through a more fine-grained resolution through an ensemble ordering, \ac{pd} assigns more than half the samples to the category of the greatest difficulty even in this scenario.
%while at the same time clearly limiting its expressiveness.
The loss- or margin-based scoring functions \ac{celoss}, \ac{cvloss}, and \ac{tt}, as well as C-score, on the other hand, have (almost) completely unique difficulty values assigned to each sample -- at least in the ensemble version. 
Any non-uniqueness, especially in the single-score version, results from limited numerical precision for very small loss values. 
Therefore, they provide a clear ordering in terms of their estimation of difficulty.

The effects of the coarse granularity on robustness are particularly reflected in  \Cref{fig:seed_aggregation_correlation}. 
Here, \ac{fit}, \ac{cumacc}, and \ac{pd} all show clearly higher robustness to randomness than the more fine-grained \ac{celoss}.

\begin{table}[t]
\centering
  \caption{
  Granularity of CIFAR-10 and DCASE2020 \ac{sf} difficulty distributions for single seed ordering and ensemble orderings.
  We report the number of unique difficulty values and the maximum number of samples assigned to a single difficulty value (bin).
  }
  \label{tab:difficulty_distributions}
  \begin{tabular}{lrr|rr}
    \toprule
    \textbf{Ensemble Size} & \multicolumn{2}{c}{\textbf{1}} & \multicolumn{2}{|c}{\textbf{6}} \\
    \midrule
    \textbf{Scoring Function} & \textbf{Unique} & \textbf{Max Bin} & \textbf{Unique} & \textbf{Max Bin} \\
    \midrule
    \multicolumn{5}{l}{\textbf{CIFAR-10}} \\
    \midrule
    CELoss & 17\,424 & 5164 & 49\,844 & 10 \\
    CVLoss & 32\,872 & 768 & 50\,000 & 1 \\
    CumAcc & 33 & 18\,279 & 157 & 4033 \\
    FIT & 46 & 18\,279 & 34\,286 & 3824 \\
    PD & 20 & 21\,135 & 196 & 10\,915 \\
    TT & 50\,000 & 1 & 50\,000 & 1 \\
    C-score & -- & -- & 50\,000 & 1 \\
    \midrule
    \multicolumn{5}{l}{\textbf{DCASE2020}} \\
    \midrule
    CELoss & 8402 & 470 & 13962 & 1 \\
    CVLoss & 13262 & 22 & 13962 & 1 \\
    CumAcc & 32 & 1889 & 151 & 670 \\
    FIT & 48 & 1889 & 12384 & 670 \\
    PD & 21 & 4119 & 203 & 2881 \\
    TT\tablefootnote{The ensemble ordering consists of three different scores, as \ac{tt} only utilizes pre-trained models.} & 13962 & 1 & 13962 & 1 \\
    \bottomrule
  \end{tabular}
\end{table}

\subsubsection{Agreement of Difficulty Notion}
\label{sec:exp_scoring_similarity}
Given the previous insights on the characteristics of \acp{sf} on an individual level, we now extend our analysis to the similarity of difficulty notions \emph{across} \acp{sf}. 
For this, we calculate pairwise Spearman rank correlation coefficients of the different \acp{sf} based on their ensembles \wrt seed, model architecture (including initialisation), and optimisation routine, with all results presented in the appendix (\cf \Cref{fig:seed_agreement}, \Cref{fig:model_agreement}, and \Cref{fig:optim_agreement}).
Evidently, almost all \acp{sf} share a very similar notion of difficulty.
Only \ac{pd} shows an uncharacteristically low, yet still moderate correlation to all other \acp{sf} of above 0.5 in almost all cases.
This indicates some limitations of the approach, likely due to the coarse granularity as well as the applied \ac{gap}.
With all other approaches agreeing to more than 70\,\% in all but one case, their ensemble-based difficulty orderings generally seem to be good estimators for model-based \ac{sd} in the context of \ac{cl}. % -- at least when hyperparameters are kept fixed.

\Cref{tab:scoring_function_robustness} summarises the \ac{sf} agreement per ensemble type as an average of the pairwise agreements, as, \eg displayed in \Cref{fig:seed_agreement}. 
Interestingly, the agreement across \acp{sf} is even higher than most agreements within the same \ac{sf} with different random seeds. % , as reported in \Cref{tab:scoring_function_robustness}.
While this might seem surprising at first, the reasons for this may lie in the fact that most approaches are based on identical model trainings. 
For instance, a likely scenario might be that a model is confronted with a certain sample early on during training and is able to learn it quickly. Consequentially, the sample might be correctly classified in all epochs, making it both `easy' in the context of \ac{cumacc} and \ac{fit}, and having a persistently low loss value, making it `easy' in the context of \ac{celoss}.

A high correlation of \acp{sf} is also observed when ensembles are constructed \wrt varying model architectures and optimisation routines, as summarised in \Cref{fig:model_agreement} and \Cref{fig:optim_agreement}.
The mean correlations across all ensembles are reported in \Cref{tab:scoring_function_agreement}.

% Agreement of ensembles (with 6 different variations for each \ac{sf}).
% C-score is the same for seed, model, optimiser variations as these are the publicly available ones.

\begin{table}[t]
\centering
  \caption{
  Mean \ac{sf} agreement for CIFAR-10 and DCASE2020.
  Reported is the mean pairwise agreement of ensemble \acp{sf} under variation of either random seeds, model architectures, or optimiser-learning rate combinations.
  }
  \label{tab:scoring_function_agreement}
  \begin{tabular}{lrr}
    \toprule
    \textbf{Variation} & \textbf{CIFAR-10} & \textbf{DCASE2020}\\
    \midrule
    Seed & .724 $\pm$ .097 & .689 $\pm$ .129\\
    Model & .673 $\pm$ .085 & .652 $\pm$ .132\\
    Optim + LR& .732 $\pm$ .112 & .724 $\pm$ .080\\
    \bottomrule
  \end{tabular}
\end{table}

\subsection{Curriculum Learning}
\label{sec:exp_curriculum}
Despite the initial motivation through \ac{cl}, we have so far only investigated \acp{sf} in isolation regarding their alignment of difficulty orderings.
This section focuses on the practical implications of \ac{sd} orderings for \ac{cl}.
We perform experiments as described in \Cref{sec:meth_pacing_functions}, based on the ensemble \ac{sd} orderings obtained from \Cref{sec:exp_scoring_similarity}.
This approach aims to provide further insights into how the performance of a model changes when we alter the data it is exposed to based on different scoring functions.

\subsubsection{Different orderings and pacing functions}
\label{sec:exp_curriculum_cl-acl-rcl}

% pacing experiments setup
% SF: CELoss, CumAcc, PD (all seed-based), TT (model-based), C-score (only CIFAR), Random (only for CL) (ensemble versions for all)
% PF: Log, Root, Linear, Exp
% final iteration (a): 0.5, 0.8
% intial size (b): 0.2
% seeds: 100, 101, 102
% sampling mode: balanced (class-balanced training subsets as long as possible)
% training setup: EfficientNet-B0 with best configuration for each dataset
% training time: equivalent to 50 epochs

% runs CIFAR: (5 CL + 5 ACL + 1 RCL) SF * 4 PF * 2 a * 3 seed = 264
% runs DCASE: (4 CL + 4 ACL + 1 RCL) SF * 4 PF * 2 a * 3 seed = 216

We start the experimental investigation of \ac{cl} with very coarse differences in the training setup.
While we did notice differences in the agreement of difficulty values across \acp{sf} and training settings, the results of \Cref{sec:exp_scoring} provide enough evidence to assume that either of the \acp{sf} is a suitable estimator for a distinction between the easiest and the most difficult samples within a dataset.
Moreover, to limit computational demands, we exclude the \ac{cvloss} and \ac{fit} \acp{sf} from the subsequent performance experiments, as a correlation above 80\,\% is noted to \ac{cumacc} for seed-based ensembles across both datasets.

In our first experiments, we evaluate how model performance is impacted when examples are presented in the intended curriculum ordering (easy-to-hard; \ac{cl}), the reverse ordering (hard-to-easy; \ac{acl}), and a completely random ordering (\ac{rcl}).

For a robust evaluation across the different orderings, we choose the seed-based ensembles for all \acp{sf}.
For training, we select the EfficientNet-B0 architecture in combination with the best baseline training setup, as preliminary experiments indicated that this configuration also shows high performance in combination with different pacing functions.
We evaluate each \ac{sf} across four \acp{pf} (logarithmic, root, linear, exponential), starting with an initial training dataset size of $b=20\,\%$ and a saturation on the full dataset after $a=50\,\%$ and $a=80\,\%$ of the training iterations.
Following the approach of~\cite{score-ce-loss-hacont}, we incorporate new training samples in a class-balanced manner, such that the training subsets remain balanced throughout training, assuming the full dataset is balanced.
Analogous to~\cite{misc-when-do-curricula-work}, we also delay the introduction of new samples until all examples in the current training subset have been used at least once.
We train each model for a number of steps equivalent to 50 epochs of the full dataset and replicate the experiments across three different seeds, averaging the performance to ensure robustness.
Overall, this leads to a total of 264\footnote{As we utilise the publicly available C-score difficulty values, we only apply the \ac{sf} to the CIFAR-10 dataset.} \ac{cl} experiments for the CIFAR-10 and 216 for the DCASE2020 dataset.

\begin{figure}[t]
    \centering
    \includegraphics[width=\columnwidth]{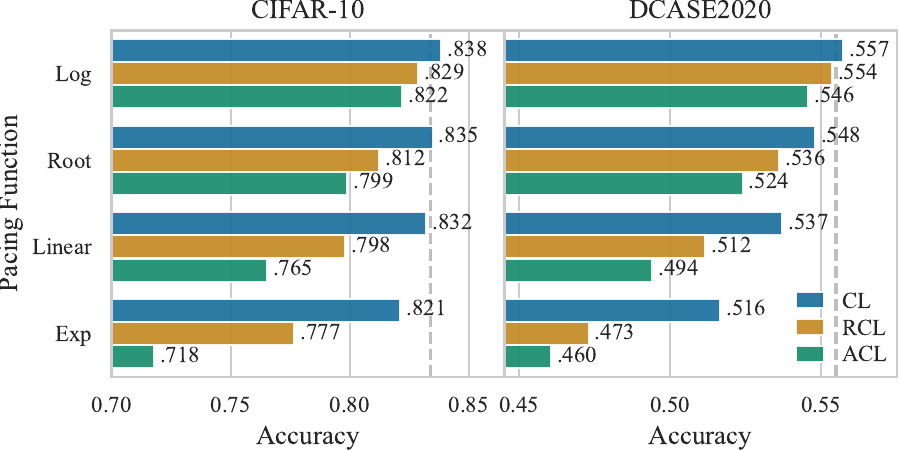}
    \caption{
    Mean \ac{pf} performance on CIFAR-10 and DCASE2020, averaged across \acp{sf}, saturation fractions, and three seeds.
    Each bar represents the average performance for each \ac{pf} and curriculum ordering.
    The grey dashed vertical lines indicate the baseline performance, averaged across the 15 random seeds.
    }
    \label{fig:pacing_fn_performance}
\end{figure}

The results of our experiments -- averaged over \acp{sf} -- as well as the baseline performance without \ac{cl} are summarised in \Cref{fig:pacing_fn_performance}.
% Additionally, we report the averaged across 15 random seeds, of 0.834 for CIFAR-10 and 0.555 for DCASE2020.
For each ordering strategy, \acp{pf} are sorted from top to bottom by decreasing saturation speed, \ie pacing functions at the top quickly introduce new samples to the training set, while those at the bottom initially maintain the number of samples close to the initial training set.
Firstly, we observe a clear trend towards better performance obtained from quickly saturating \acp{pf}, which is in line with the findings in~\cite{misc-when-do-curricula-work}.
Only logarithmic \acp{pf} combined with a \ac{cl} ordering, but not with a \ac{rcl} or \ac{acl} ordering, are able to marginally surpass the baseline on both datasets.

Our experiments further show a clear performance advantage of \ac{cl} over \ac{acl}, with \ac{rcl} scoring in between \ac{cl} and \ac{acl}. 
Despite being less prominent for the quickly saturating \acp{pf}, this difference steadily increases with decreasing saturation speed. 
In the case of the slowly saturating \acp{pf} in particular, models are trained for a long time with only the easiest (\ac{cl}), the most difficult (\ac{acl}), or random samples (\ac{rcl}), respectively. 
While most curriculum orderings and \ac{pf} lead to detrimental performance to some degree, which can likely be attributed to an effect of overfitting from which the model struggles to recover at the later stages of the training, the effects are the strongest for \ac{acl} and the weakest for \ac{cl}.
We conclude that model training is clearly negatively impacted if confronted with difficult samples first.
This finding underlines the importance of the data that models are exposed to early on. %and, to the best of the authors knowledge, has not been reported in literature to this extent.
It provides evidence supporting the concepts behind \ac{cl} and adds to the validity of model-based \acp{sf} for \ac{sd} estimation.

% Despite the intuitive nature of this presented result 
% Best \ac{cl}, \ac{acl}, and \ac{rcl} runs, averaged across 3 seeds.
% Compared to the best single seed baseline, best baseline over 3 and 6 seeds.

% All of these use seed-based ensembles for the scoring functions and EfficientNet-B0 with the best baseline training setup.

% Epochs are adjusted to the number of training iterations in the case of CL-based experiments, e.g. both performed the same number of steps.

\subsubsection{Benefits of Robust Scoring Functions for Curriculum Learning}
\label{sec:exp_curriculum_robustness}
Despite the apparent importance of data ordering for \ac{cl} settings in the extreme cases of easy-to-hard, hard-to-easy, and random difficulty orderings, we further aim to investigate the more subtle impacts \ac{sf}-based orderings have on \ac{cl}.
To further analyse this question, we revisit the robustness experiments of \acp{sf} from \Cref{sec:exp_scoring_ensemble}.
We hypothesise that ensemble \acp{sf} of larger size -- being more robust to randomness -- should have a clearer notion of difficulty, which in turn should lead to a more reliable difficulty ordering and thus have benefits in a \ac{cl} setting.

\begin{figure}[t]
    \centering
    \includegraphics[width=\columnwidth]{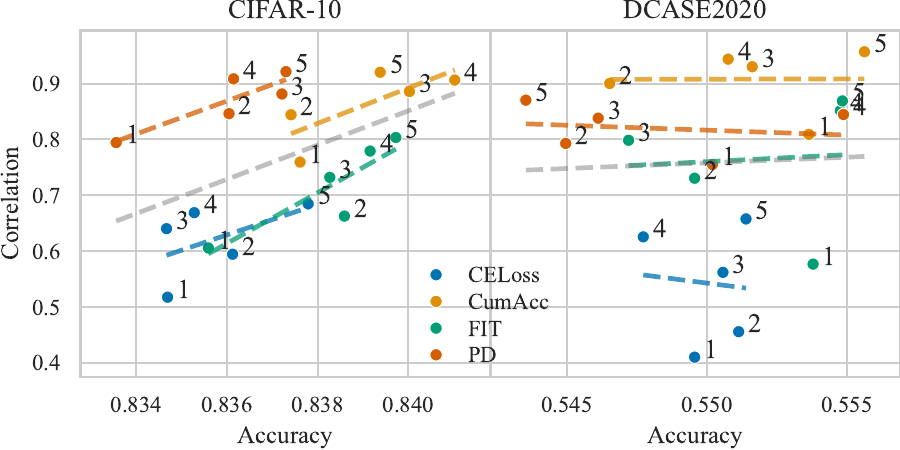}
    \caption{
    Comparison of \ac{sf} robustness and \ac{cl} performance for CIFAR-10 and DCASE2020.
    For each scoring function, we evaluate ensemble orderings of different ensemble sizes, noted as a number next to each point.
    The $y$-axis represents the pairwise correlation across the ensembles of the respective ensemble size (cf. $x$-axis in \Cref{fig:seed_aggregation_correlation}) as an indicator of \ac{sf} robustness.
    The $x$-axis displays the average accuracy of \ac{cl} experiments based on the corresponding ensemble orderings.
    Coloured and grey dashed lines are linear least-squared-error fits per \ac{sf} and across all \acp{sf}, respectively.
    The slope of the lines indicates whether trends of higher \ac{sf} robustness and higher \ac{cl} performance exist.
    }
    \label{fig:seed_aggregation_performance}
\end{figure}

In order to investigate this hypothesis, we base our \ac{cl} experiments on the ensemble orderings previously investigated in \Cref{sec:exp_scoring_ensemble}. 
We once again use three different orderings per ensemble size and three different random seeds in the same setting as \Cref{sec:exp_curriculum_cl-acl-rcl}, resulting in overall 60 curriculum experiments per \ac{sf} for each dataset.
\Cref{fig:seed_aggregation_performance} displays the average pairwise correlation for each ensemble size, \ie the values on the $y$-axis in \Cref{fig:seed_aggregation_correlation}, versus the average performance of the \ac{cl}-based training with the same ensemble size. 
% Given the limited agreement of \ac{pd} with other \acp{sf} and the other previously iterated limitations, we exclude \ac{pd} from the discussion.

At this point, results for CIFAR-10 and DCASE2020 seem to diverge considerably.
The outcomes for CIFAR-10 clearly support our hypothesis both within and across \acp{sf} (correlation across all points), with high \acp{pcc} between accuracy and correlation of 0.760 and 0.498, respectively.
The first value is calculated as the average correlation for each \ac{sf} separately (macro \ac{pcc}), while the second value is based on the correlation of all points independent of the concrete \acp{sf} (micro \ac{pcc}).
This aligns with our expectation that a more robust \ac{sd} ordering will result in improved performance when used as the starting point for \ac{cl}.

DCASE2020, however, adds more ambiguity to the discussion: both the average correlation within and across \acp{sf} show close to no correlation of -0.047 and 0.045, respectively.
In this case, a higher agreement on the \ac{sd} did not translate to improved performance when this ordering was used for \ac{cl}, indicating that other factors might influence \ac{cl} behaviour.

While we find clear evidence supporting the benefit of more robust \acp{sf} for \ac{cl} on the CIFAR-10 dataset, no such evidence can be reported for DCASE2020, suggesting further investigation into how \ac{sf} robustness and other factors contribute to \ac{cl} performance.

\subsubsection{Scoring Function and Performance}
\label{sec:exp_curriculum_performance}
Having compared the effects of different orderings within the \ac{cl} setting, the next aspect we aim to investigate is whether the \ac{cl} methodology is in any way superior to standard \ac{dl} training beyond the limited training time scenario reported in \cite{misc-when-do-curricula-work}.
To test this, we select the 264 \ac{cl} experiments for the CIFAR-10 and 216 for the DCASE2020 dataset from \ref{sec:exp_curriculum_cl-acl-rcl}.
We report the best-performing combination for easy-to-hard (\ac{cl}), hard-to-easy (\ac{acl}), and the random ordering (\ac{rcl}) averaged across three seeds, respectively.
For comparison, we use the 15 models from \ref{sec:exp_scoring_ensemble}, each replicating the EfficientNet-B0 reference configuration across different random seeds, as baselines.
We report the performance of the single best model (B1), the average performance of the best three models (B3), the best five models (B5), and all baselines (B15).
This setup allows us to investigate whether \ac{cl}, \ac{rcl}, or \ac{acl} can offer performance benefits over standard training across both datasets in order to challenge the findings of Wu \etal \cite{misc-when-do-curricula-work}, which reported no improvement over the baseline.
Despite the differing number of \ac{cl} and baseline experiments, the comparison aims to assess whether significant performance improvements can be achieved through \ac{cl}-based training.
However, the choice of the best-performing \ac{cl}-based configurations introduces a selection bias, as they represent the best training setup for \ac{cl}, while the baselines are derived from a fixed training setup averaged across multiple (best-performing) random seeds.

%It is important to note that in the comparison to the best baseline runs (B1, B3, B5, and B15), we tend to put the \ac{cl} approaches at a disadvantage, as the hypothesis to test is that \ac{cl} improves the performance.
%However, the selection of the best performing \ac{cl} configuration again poses a selection bias.
%emphasize more the difference instead of picking the disadvantage
%It is important to note the limitations of the comparison is TBL5.
%The baseline runs (in particular ... B1) contain a selection bias of the best performing random seeds.
%For \ac{cl} however, we select the best performing model overall which is then averaged across three seeds on the other hand.
%However the results aim to show whether a performance increase through \ac{cl} is ACTUALLY EVEN achievable (ref when do curricula work paper).
%(In contrast to what is reported in when do curricula work)
% This comparison allows us to assess whether \ac{cl}-based approaches can even provide a plug-and-play improvement and surpass the baseline, given the significant advantage of a larger number of training combinations.

\Cref{tab:performance} provides an overview of the performance experiments.
The results, however, are inconclusive.
For CIFAR-10, the best performance is indeed achieved by a \ac{cl}-based configuration using the computation-heavy C-score ordering.
However, the next best performances are achieved by all baselines, followed by \ac{rcl} and, finally, \ac{acl} showing the worst performance.
For DCASE2020, the best \ac{cl} setting cannot reach the performance of the best baseline run but outperforms the average performance from the best 3 to 15 baselines, with also \ac{acl} and \ac{rcl} outperforming the average overall baseline runs.
A clear plug-and-play improvement of \ac{cl} over standard training across datasets can therefore not be concluded, which falls in line with the results from~\cite{misc-when-do-curricula-work}.
Despite the apparent limitations of this analysis due to the asymmetric selection of comparison models, it seems that a good \ac{cl} setting does not have substantially beneficial effects on the performance over the standard training baselines.

\begin{table}[t]
\centering
  \caption{
  Performance comparison between the averages of the best standard training baselines (abbreviated with B) and the best training settings for \ac{cl}, \ac{rcl}, and \ac{acl} averaged across three seeds.
  }
  \label{tab:performance}
  \begin{tabular}{llrllr}
    \toprule
    \multicolumn{3}{l}{\textbf{CIFAR-10}} & \multicolumn{3}{l}{\textbf{DCASE2020}} \\
    \midrule
    \textbf{SF} & \textbf{Type} & \textbf{Accuracy} & \textbf{SF} & \textbf{Type} & \textbf{Accuracy} \\
    C-score & CL & .844 & -- & B1 & .583 \\
    -- & B1 & .839 & TT & CL & .577\\
    -- & B3 & .839 & -- & B3 & .576 \\
    -- & B5 & .838 & -- & B5 & .571 \\
    -- & B15 & .834 & CELoss & ACL & .560 \\
    Random & RCL & .829 & Random & RCL & .558 \\
    CELoss & ACL & .829 & -- & B15 & .555 \\
    \bottomrule
  \end{tabular}
\end{table}

\subsubsection{Late Fusion Performance}
Our last line of experiments focuses on the question of whether models trained with a focus on different samples arrive at states that are complementary to each other.
We investigate this through the lens of late fusion, which is performed by averaging the predicted class probabilities of the respective models after the softmax layer.
In this context, we expect higher performance gains for combinations that show higher independence in their voting~\cite{kuncheva2003limits}, which we interpret as more disconnected concepts across models. 
Therefore, we select the best-performing model trained with easy-to-hard, hard-to-easy, and random orderings alongside the baselines from \ref{sec:exp_curriculum_performance}.
We denote the fusion of the top two baselines as B2, the fusion of the top three baselines as B3, and so forth.

\begin{figure}[t]
    \centering
    \includegraphics[width=\columnwidth]{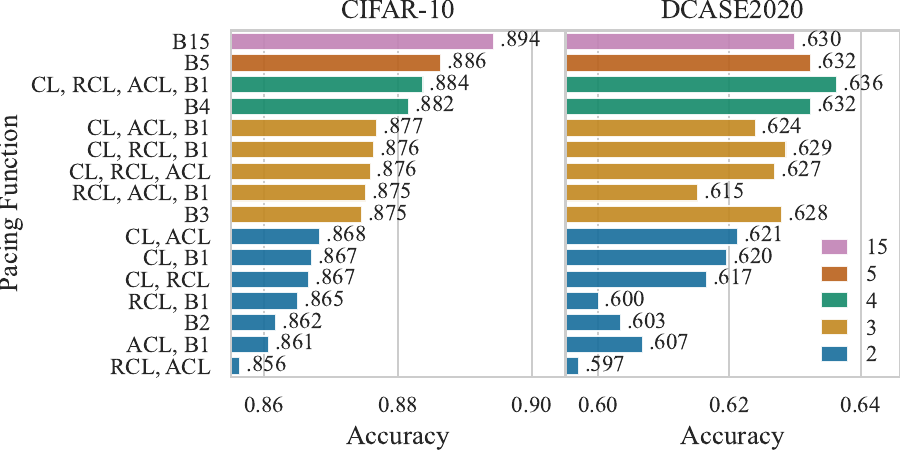}
    \caption{
    Late fusion results for combinations of curriculum (\ac{cl}), random curriculum (\ac{rcl}), and anti-curriculum learning (\ac{acl}), as well as the best baselines abbreviated with B; for instance, B4 represents the fusion of the best 4 baseline runs.
    }
    \label{fig:late_fusion}
    % we also have a version with sizes 2-4 if needed (combined_fusion.pdf)
\end{figure}

\Cref{fig:late_fusion} provides an overview of the performance obtained through late fusion of differently trained model states.
Across all experiments, we notice a clear performance increase when fusing multiple models, as would be expected from the literature~\cite{dietterich2000ensemble, fort2019deep}.
For CIFAR-10, the highest performance across all experiments is achieved by the fusion of all baseline models.
In contrast, for the DCASE2020 dataset, the highest performance is obtained from the fusion of \ac{cl}, \ac{acl}, \ac{rcl}, and the best baseline.
Among fusions of two, three, and four models, we find that combinations including any of the \ac{cl}-based orderings tend to outperform baseline fusions of the same size across both datasets, especially if \ac{cl} and \ac{acl} are part of the combination.
This is particularly interesting as \ac{acl} and \ac{rcl} have a low performance compared to the best baselines (\cf\Cref{tab:performance}).
The fact that the fusion of \ac{cl} and \ac{acl} performs better suggests that these opposing difficulty orderings may complement each other, indicating an exploitable difference in the learnt concepts between \ac{cl} and \ac{acl}.

\section{Conclusion}
\label{sec:conc}
In this contribution, we extensively explored the robustness and alignment of sample difficulty estimation through scoring functions and their implications for curriculum learning on benchmark computer vision and computer audition datasets.
We discovered that model-based scoring functions are almost to an equal level impacted by randomness in terms of parameter initialisation and dataset order, as they are by the choice of hyperparameters in the respective training setting. 
Robustness towards randomness, however, was shown to be effectively increased through ensemble scoring functions.
A generally high agreement across different ensembles of scoring functions based on various concepts could be observed as long as their difficulty predictions allowed for a fine-grained estimation of difficulty and, correspondingly, a clear profile of the difficulty ordering.

In the context of curriculum learning, we first showed a clear benefit of easy-to-hard over hard-to-easy difficulty orderings, in particular for pacing functions that slowly saturate on the full dataset.
We further found evidence that more robust scoring functions benefit curriculum learning, likely through a more robust notion of difficulty.
As we could not show that curriculum learning has a general advantage over traditional deep learning training, we are left to hypothesise that many of the reported benefits in the literature are not universally generalisable.
Nevertheless, models trained with opposing difficulty orderings revealed beneficial properties for late fusion.
This leads to the conclusion that different, complementary concepts can be acquired when models learn samples easy-to-hard versus hard-to-easy.
Future work could expand our findings by investigating the characteristics of different samples and how those impact their difficulty ranking.

\section*{Acknowledgment}
This work has received funding from the DFG's Reinhart Koselleck project No.~442218748 (AUDI0NOMOUS).

\clearpage

\appendix

\section*{Baselines}

\begin{table}[h]
\centering
  \caption{
  Grid search hyperparameters utilised to establish the baseline performance.
  Each model was trained 50 epochs, with the final model selection based on the best-performing state on the respective validation set.
  }
  \label{tab:grid_search}
  \begin{tabular}{ll}
    \toprule
    \textbf{Hyperparameter} & \textbf{Values}\\
    \midrule
    Architecture &  EfficientNet-B0, -B4, ResNet50, CNN10, CNN14\tablefootnote{EfficientNet-B4 and ResNet50 only are only trained for CIFAR-10, CNN10 and CNN14 only for DCASE2020}\\
    Initialisation & Random, Pre-trained\tablefootnote{EfficientNets and ResNet50 are pre-trained on ImageNet, CNN10 and CNN14 are pre-trained on AudioSet}\\
    Optimiser & Adam~\cite{optim-adam}, SGD\tablefootnote{with momentum (0.9)}~\cite{optim-sgd-momentum}, SAM\tablefootnote{with SGD momentum (0.9)}~\cite{optim-sam}\\
    Learning rate & .01, .001, .0001\\
    Batch size & 16\\
    Epochs & 50\\
    Random Seed & 1\\
    \bottomrule
  \end{tabular}
\end{table}

\begin{table}[h]
\centering
  \caption{
  Best baseline performance of each model on CIFAR-10 and DCASE2020 with variations in optimisers and learning rates as shown in \Cref{tab:grid_search}.
  The suffix -T indicates pre-training (on ImageNet for CIFAR-10 and AudioSet for DCASE2020).
  All models were trained for 50 epochs, with final selection based on the best validation performance.
  }
  \label{tab:performance_baselines}
  \begin{tabular}{lrrr}
    \toprule
    \textbf{Model} & \textbf{Optimiser} & \textbf{Learning Rate} & \textbf{Accuracy}\\
    \midrule
    \multicolumn{4}{l}{\textbf{CIFAR-10}} \\
    \midrule
    ResNet-50-T	& SAM & .001 & .949 \\
    EfficientNet-B4-T & SAM & .001 & .945 \\
    EfficientNet-B0-T & SAM & .001 & .936 \\
    EfficientNet-B4 & Adam & .001 & .848 \\
    EfficientNet-B0 & Adam & .001 & .835 \\
    ResNet-50 & Adam & .001	& .813 \\
    \midrule
    \multicolumn{4}{l}{\textbf{DCASE2020}} \\
    \midrule
    CNN14-T & Adam & .0001 & .678 \\
    CNN10-T & SAM & .01 & .653 \\
    EfficientNet-B0-T & SAM & .01 & .644 \\
    CNN10 & SAM & .001 & .609 \\
    CNN14 & SAM & .001 & .595 \\
    EfficientNet-B0 & SAM & .01 & .583 \\
    \bottomrule
  \end{tabular}
\end{table}

\newpage
\section*{Agreement of Difficulty Notion}

\begin{table}[h]
\centering
  \caption{
  Overview of variations in random seed, model architecture (with EfficientNets abbreviated to B0 and B4) and initialisation, and optimiser-learning rate combinations used for \ac{sf} calculation.
  The suffix -T indicates pre-training (on ImageNet for CIFAR-10 and AudioSet for DCASE2020) across model architectures.
  Each training setting is varied independently, with other parameters fixed to the first entry in each row, based on the EfficientNet-B0 reference configuration.
  }
  \label{tab:scoring_function_variations}
  \begin{tabular}{lp{3cm}p{3cm}}
    \toprule
    \textbf{Variations} & \textbf{CIFAR-10} & \textbf{DCASE2020}\\
    \midrule
    Seed & 1, 2, 3, 4, 5, 6 & 1, 2, 3, 4, 5, 6 \\
    Model & B0, B0-T, B4, B4-T, ResNet50, ResNet50-T & B0, B0-T, CNN10, CNN10-T, CNN14\tablefootnote{The learning rate for CNN14 was reduced, as the model did not converge with all other parameters fixed to the first entry in each row.}, CNN14-T \\
    Optim + LR & Adam, SAM, SGD (all with .001, .01) & SAM, Adam, SGD (all with .01, .001)\\
    \bottomrule
    \end{tabular}
\end{table}

\begin{figure}[h]
    \centering
    \subfloat[CIFAR-10]{\includegraphics[width=0.98\columnwidth]{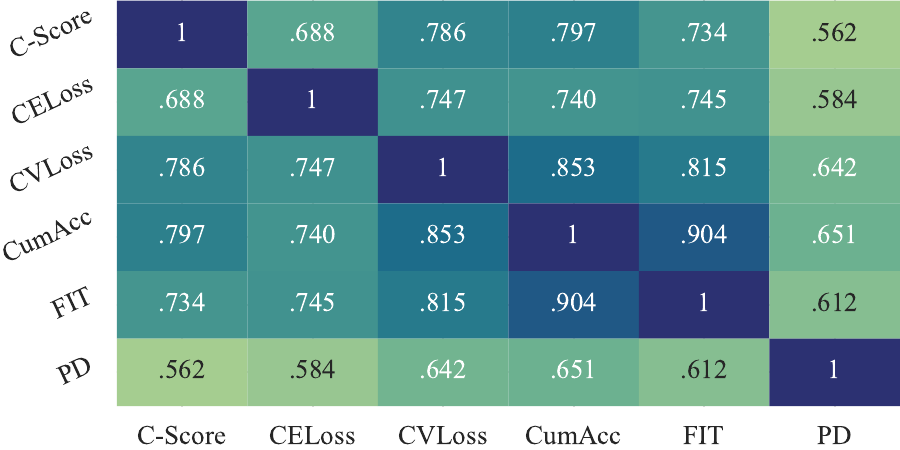}
    %\label{fig:cifar10_seed_agreement}
    }\\
    \subfloat[DCASE2020]{\includegraphics[width=0.98\columnwidth]{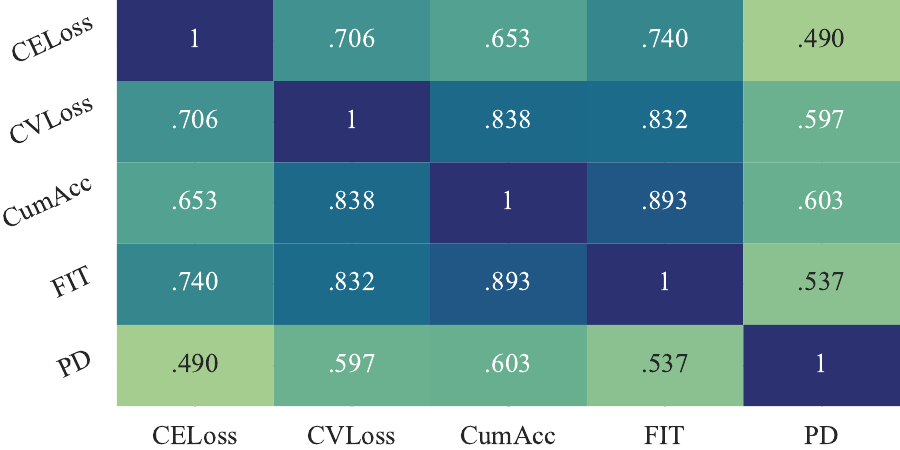}
    %\label{fig:dcase2020_seed_agreement}
    }
    \caption{
    Agreement of different scoring functions with varying random seeds.
    Displayed is the pairwise Spearman correlation of the respective ensemble orderings of ensemble size six.
    The individual orderings building up the ensemble are obtained from the reference configuration and five additional variations of the random seed.
    }
    \label{fig:seed_agreement}
\end{figure}

\begin{figure}[h]
    \centering
    \subfloat[CIFAR-10]{\includegraphics[width=0.98\columnwidth]{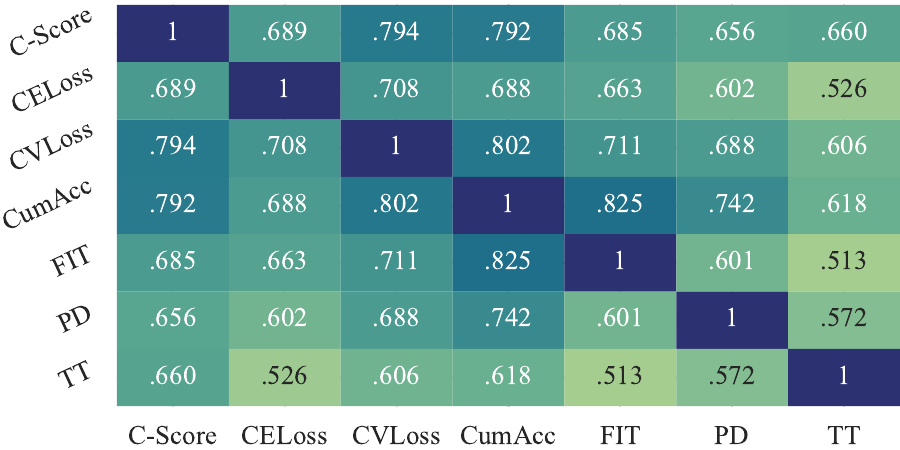}
    %\label{fig:cifar10_model_agreement}
    }\\
    \subfloat[DCASE2020]{\includegraphics[width=0.98\columnwidth]{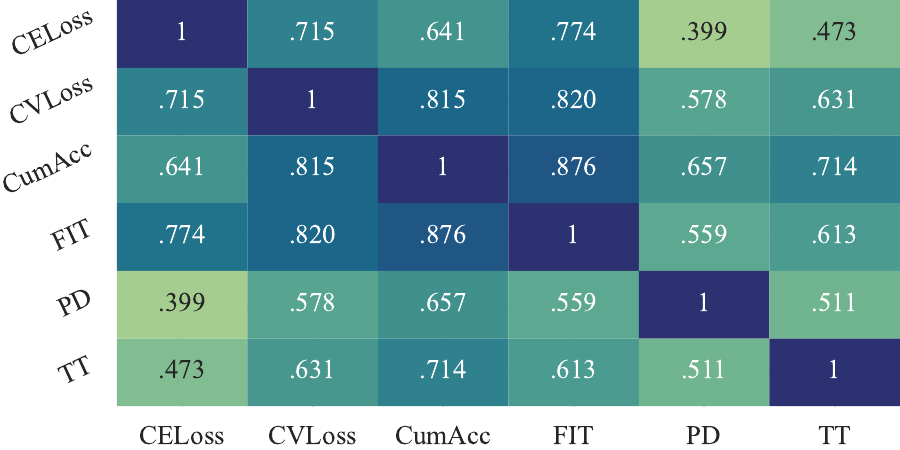}
    %\label{fig:dcase2020_model_agreement}
    }
    \caption{
    Agreement of different scoring functions with varying models.
    Displayed is the pairwise Spearman correlation of the respective ensemble orderings of ensemble size six.
    The individual orderings building up the ensemble are obtained from the reference configuration and five additional variations of the model architecture and initialisation.
    }
    \label{fig:model_agreement}
\end{figure}

\begin{figure}[h]
    \centering
    \subfloat[CIFAR-10]{\includegraphics[width=0.98\columnwidth]{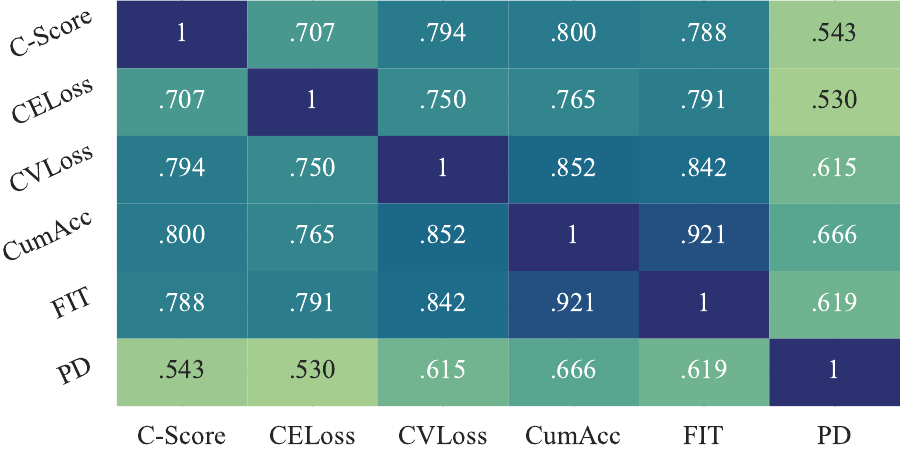}
    %\label{fig:cifar10_optim_agreement}
    }\\
    \subfloat[DCASE2020]{\includegraphics[width=0.98\columnwidth]{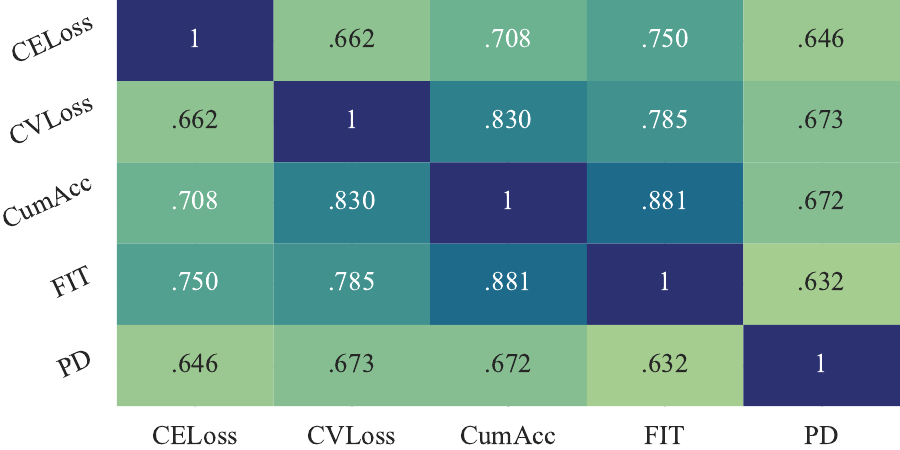}
    %\label{fig:dcase2020_optim_agreement}
    }
    \caption{
    Agreement of different scoring functions with varying optimiser and learning rate combinations.
    Displayed is the pairwise Spearman correlation of the respective ensemble orderings of ensemble size six.
    The individual orderings building up the ensemble are obtained from the reference configuration, two additional optimiser variations, and the two best-performing learning rates for each optimiser.
    }
    \label{fig:optim_agreement}
\end{figure}

%TODO: check new pages
\newpage % or \clearpage
\bibliographystyle{IEEEtran}
\bibliography{references}
%\newpage

\begin{IEEEbiography}[{\includegraphics[width=1in,height=1.25in,clip,keepaspectratio]{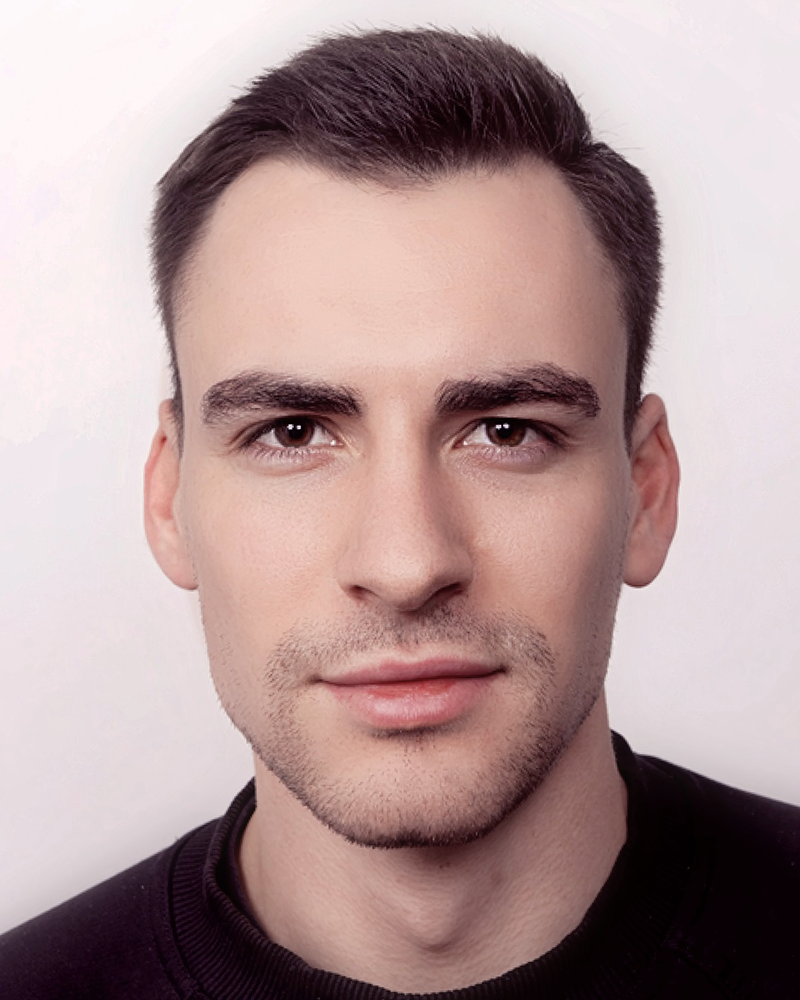}}]{Simon Rampp}
received his Bachelor of Science in Computer Science from the University of Augsburg in 2022 and his Master of Science from the same university in 2024.
He is currently conducting research as a guest at the chair of Health Informatics, Technical University of Munich.
His work focuses on deep learning for computer vision and computer audition, as well as understanding sample difficulty for curriculum learning.

\end{IEEEbiography}
%\vspace{-250pt}
\begin{IEEEbiography}[{\includegraphics[width=1in,height=1.25in,clip,keepaspectratio]{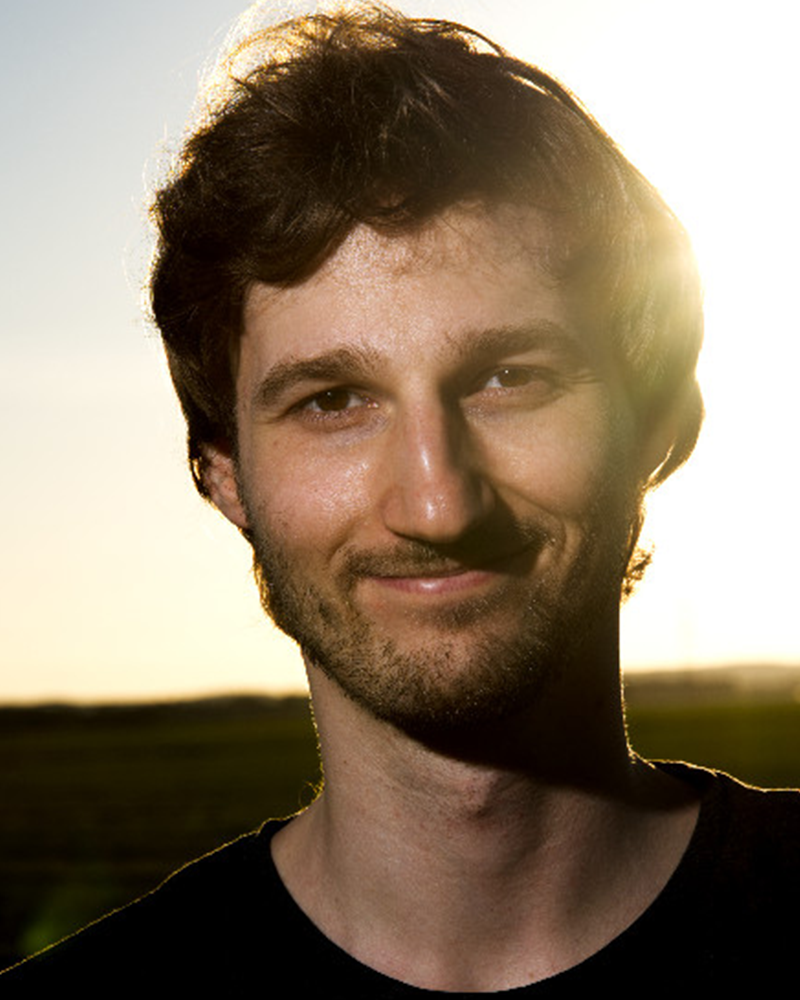}}]{Manuel Milling}
received his Bachelor of Science in Physics and in Computer Science from the University of Augsburg in 2014 and 2015, respectively, and his Master of Science in Physics from the same university in 2018.
He is currently a PhD candidate in Computer Science at the chair of Health Informatics, Technical University of Munich.
His research interests include machine learning with, a particular focus on the core understanding and applications of deep learning methodologies.
\end{IEEEbiography}
%\vspace{-250pt}
\begin{IEEEbiography}[{\includegraphics[width=1in,height=1.25in,clip,keepaspectratio]{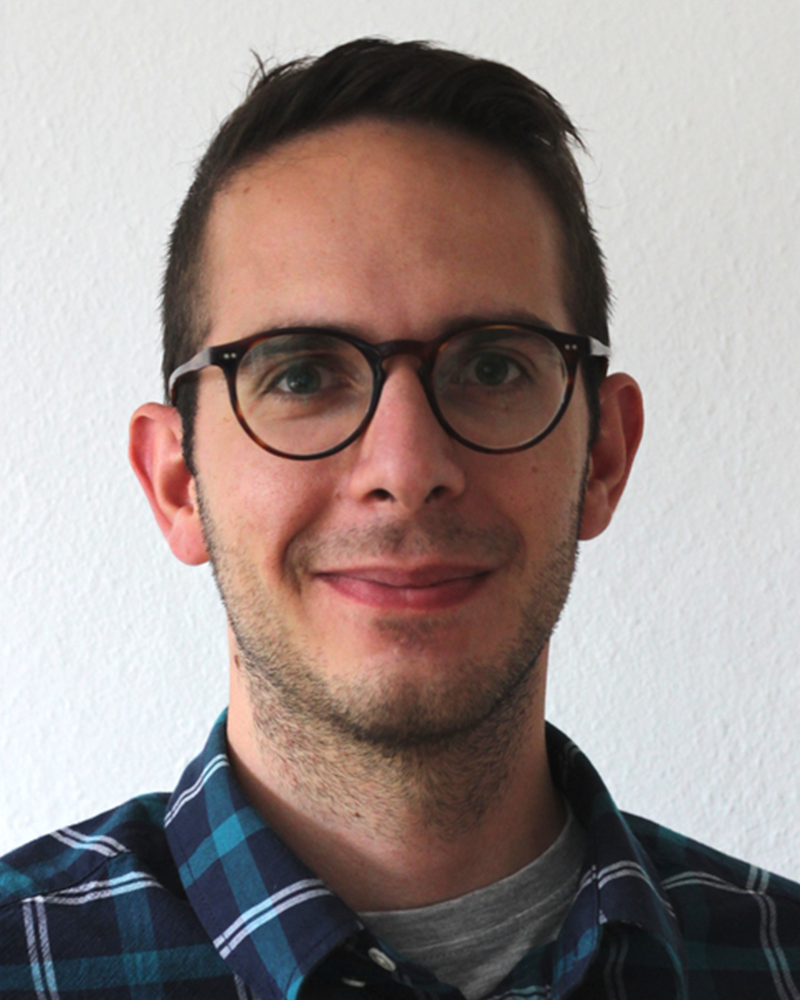}}]{Andreas Triantafyllopoulos}
received the diploma in ECE from the University of Patras, Greece, in 2017.
He is working toward the doctoral degree with the Chair of Health Informatics, Technical University of Munich.
His current focus is on deep learning methods for auditory intelligence and affective computing.
\end{IEEEbiography}
%\vspace{-250pt}
\begin{IEEEbiography}[{\includegraphics[width=1in,height=1.25in,clip,keepaspectratio]{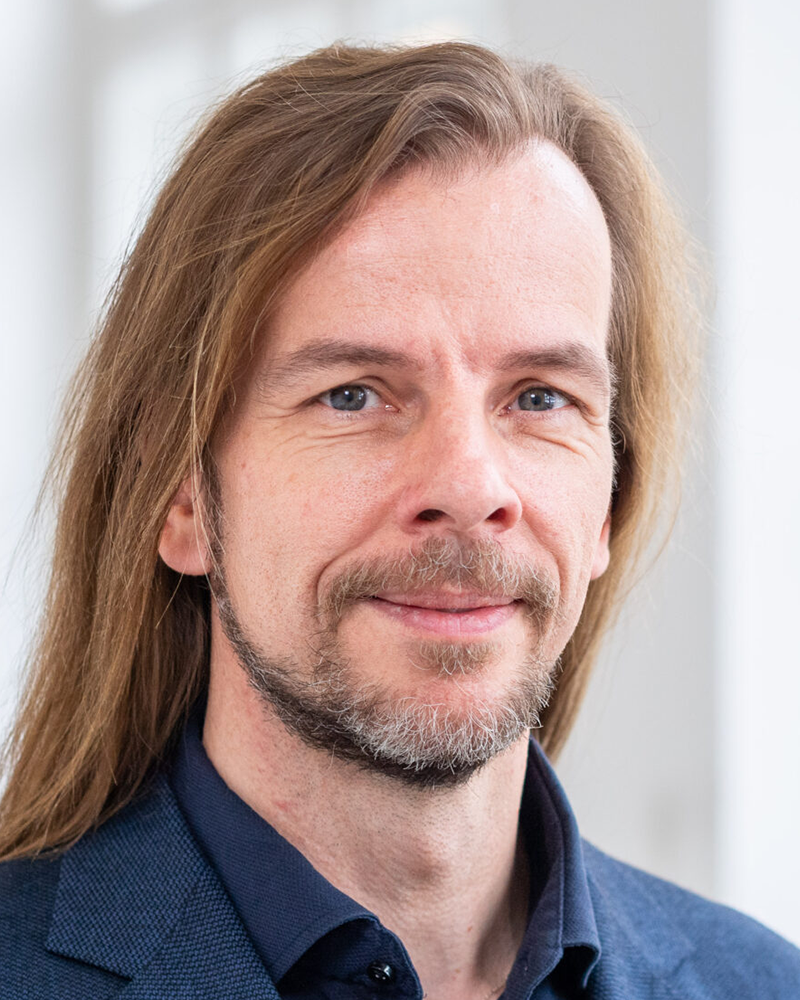}}]{Björn W.\ Schuller}
received his diploma in 1999, his doctoral degree in 2006, and his habilitation and was entitled Adjunct Teaching Professor in 2012, all in electrical engineering and information technology from TUM in Munich/Germany.
He is Full Professor of Artificial Intelligence and the Head of GLAM at Imperial College London/UK, Chair of CHI -- the Chair for Health Informatics, MRI, Technical University of Munich, Munich, Germany, amongst other Professorships and Affiliations.
He is a Fellow of the IEEE and Golden Core Awardee of the IEEE Computer Society, Fellow of the ACM, Fellow and President-Emeritus of the AAAC, Fellow of the BCS, Fellow of the ELLIS, Fellow of the ISCA, and Elected Full Member Sigma Xi.
He (co-)authored 1,400+ publications (60,000+ citations, h-index=111).
\end{IEEEbiography}

\end{document}